\documentclass[sigconf,natbib=false,nonacm,9pt,balance=false,
hyperref={hidelinks,bookmarksnumbered,unicode}]{acmart}

\usepackage{amsmath,amssymb,amsfonts}
\usepackage{graphicx}
\usepackage{multirow}
\usepackage{siunitx}
\usepackage[caption=false,font=footnotesize,labelfont=sf,textfont=sf]{subfig} 
\usepackage{url}
\usepackage{cite}
\usepackage{xcolor}
\usepackage{array}     
\usepackage{makecell}  
\usepackage{algorithmicx}  

\usepackage{algpseudocode} 
\usepackage{algorithm}
\usepackage{dblfloatfix} 
\usepackage{cuted}     
\usepackage{capt-of}   
\usepackage{booktabs}  

\usepackage{orcidlink}

\usepackage{pgfplots}
\pgfplotsset{compat=1.18}
\usepackage{xcolor}

\definecolor{ieeeblue}{RGB}{31,119,180}
\definecolor{ieeeorange}{RGB}{255,127,14}

\begin{document}

\title{Joint UAV Activation and Placement for Post-Disaster Wireless Restoration via a Hybrid Quantum-Inspired Evolutionary Framework}

\author{Fatima Azzahraa Amarcha$^{1}$, Lahcen Hassine \orcidlink{0009-0006-9878-5919}$^{2}$, Rachid Saadane \orcidlink{0000-0002-0197-8313}$^{2,3}$, Mohamed Rahouti \orcidlink{0000-0001-9701-5505}$^{4}$, Rachid Ahl Laamara \orcidlink{0000-0001-8410-9983}$^{1,5}$, Abdallah Slaoui \orcidlink{0000-0002-5284-3240}$^{1,5,*}$, and Hany S. khalifa \orcidlink{0000-0001-9462-3409}$^{6,*}$}

\affiliation{$^1$LPHE, Modeling and Simulations, Faculty of Science, Mohammed V University in Rabat, Morocco.\\
$^2$Hassania School of Public Works (EHTP), Laboratory Engineering System, SIRC Team, Casablanca, Morocco.\\
$^3$Genuis Lab, School of Management, Telecommunications and Computer Science (SUPMTI), Rabat, Morocco.
$^4$Department of Computer and Information Science, Fordham University, New York, NY, USA.\\
$^5$CPM, Centre of Physics and Mathematics, Faculty of Science, Mohammed V University in Rabat, Morocco.\\
$^6$Computer Science Department, Misr Higher Institute for Commerce and Computers, Mansoura 35511, Egypt.
\country{}
{$^*${Corresponding author: abdallah.slaoui@um5s.net.ma \& h.khalifa@metmans.edu.eg}}}

\begin{abstract}
In post-disaster environments, the failure of terrestrial communication infrastructure necessitates the rapid deployment of unmanned aerial vehicles (UAVs) as aerial base stations to restore wireless connectivity. This paper addresses the joint UAV activation-and-placement problem in continuous space, with the objective of minimizing the number of deployed UAVs while satisfying coverage and minimum-separation constraints. To solve this problem, we propose a Hybrid K-means Quantum-Inspired Evolutionary Algorithm (HKQEA) that combines K-means-guided initialization, a calibrated penalty-based feasibility objective, non-elitist evolutionary search, and a quantum-inspired learning update. Experimental results over 50 independent runs show that HKQEA attains a best fully feasible solution with 8 UAVs, while achieving average values of 98.94\% for coverage, 99.94\% for non-overlap, and 99.68\% for minimum-distance satisfaction. Comparative evaluation against standard Non-dominated Sorting Genetic Algorithm II (NSGA-II), Particle Swarm
Optimization algorithm (PSO) and an elitist variant of HKQEA further shows that the proposed method provides a more favorable balance among exploration, convergence behavior, and reliable feasibility preservation in constrained deployment problems. An illustrative procurement-level cost analysis also indicates that reducing the fleet from 10 UAVs to 8 can yield a 20\% reduction in hardware count, corresponding to a simplified savings ratio of 25\% for the studied deployment setting. These results demonstrate the potential of the proposed framework for resource-efficient post-disaster communication restoration.\\ \\
{\bf Keywords:} Unmanned Aerial Vehicles, Quantum-Inspired Evolutionary Algorithm, wireless network, K-means Initialization, Set Covering Problem.

\keywords{
{\bf Keywords:} Unmanned Aerial Vehicles, Quantum-Inspired Evolutionary Algorithm, wireless network, K-means Initialization, Set Covering Problem.
}
\end{abstract}

\maketitle
\section{Introduction}

Natural disasters such as earthquakes, floods, hurricanes, and volcanic eruptions can cause severe human and economic losses while simultaneously disrupting critical infrastructure \cite{b38}. Among the most consequential failures in the immediate aftermath of a disaster is the partial or complete breakdown of terrestrial communication networks. When connectivity is lost, affected populations may be isolated, emergency responders may face major coordination challenges, and the timely exchange of situational information may be severely constrained, ultimately delaying rescue and recovery operations \cite{b37}. Rapid restoration of wireless connectivity is therefore a critical requirement in post-disaster environments \cite{b39}.

Recent disaster events have further highlighted the operational importance of resilient emergency communications. In such settings, communication networks serve not merely as supporting infrastructure, but as a lifeline that enables survivors to report their status, responders to coordinate field activities, and authorities to prioritize resource allocation. This need has motivated growing interest in rapidly deployable communication platforms capable of temporarily replacing damaged terrestrial systems.

Unmanned aerial vehicles (UAVs) have emerged as a promising solution for this purpose. Owing to their mobility, flexibility, and comparatively low deployment cost, UAVs can be rapidly deployed as aerial base stations to provide provisional wireless coverage over affected areas \cite{b53,b40}. Beyond connectivity restoration, UAVs have also been used to support search-and-rescue and situational-awareness tasks through aerial imaging and onboard sensing technologies \cite{b41,b42,b43}. However, the coverage offered by a single UAV is inherently limited by communication range and operational endurance. As a result, practical disaster scenarios typically require the coordinated deployment of multiple UAVs.

Despite substantial progress in UAV-assisted emergency networking, an important methodological gap remains. Much of the existing literature assumes that the number of UAVs is fixed a priori and focuses primarily on placement optimization. In realistic disaster-response settings, however, the fleet size is itself a critical decision variable because excessive deployment increases logistical burden, cost, and energy usage, whereas insufficient deployment leads to incomplete service restoration. The problem is therefore not only where to place UAVs, but also how many UAVs are actually needed to ensure reliable coverage under operational constraints. When coverage requirements, overlap control, and minimum inter-UAV separation are considered jointly, the resulting deployment problem becomes highly combinatorial and can be viewed as an NP-hard constrained optimization problem with set-covering characteristics \cite{b55}.

To address optimization problems of this complexity, heuristic and metaheuristic techniques have been widely adopted because of their ability to explore large search spaces within practical computational budgets \cite{b8}. Classical approaches such as Particle Swarm Optimization algorithm (PSO) \cite{b64} and Non-dominated Sorting Genetic Algorithm II (NSGA-II) \cite{Deb2002NSGA2} have shown promise, but their performance may degrade in highly constrained deployment settings where feasibility preservation, search diversity, and convergence stability must be balanced carefully. Quantum-inspired evolutionary algorithms have attracted increasing attention in this context because probabilistic representations can enhance exploration and reduce stagnation. Nevertheless, conventional quantum-inspired approaches may still suffer from weak initial population quality, sensitivity to constraint handling, and premature convergence under elitist evolutionary strategies \cite{b56}.

Motivated by these limitations, this paper develops a hybrid quantum-inspired evolutionary framework for post-disaster UAV deployment. The objective is to minimize the number of deployed UAVs while determining spatial configurations that satisfy coverage and safety requirements. The proposed approach combines structured K-means-based initialization, calibrated penalty-based feasibility management, and a non-elitist quantum-inspired evolutionary search mechanism designed to preserve diversity and support stable convergence in constrained search spaces.

The main contributions of this work are summarized as follows:
\begin{itemize}
    \item We formulate post-disaster UAV deployment as a constrained optimization problem that jointly addresses fleet-size minimization and spatial deployment under coverage and separation requirements.
    
    \item We develop a hybrid K-means Quantum-Inspired Evolutionary Algorithm (HKQEA) that integrates structured initialization, non-elitist evolutionary search, and a quantum-inspired update mechanism to improve exploration and reduce premature convergence.
    
    \item We introduce a calibrated penalty-based constraint-handling strategy and examine its effectiveness relative to alternative feasibility-management approaches within the proposed optimization framework.
    
    \item We evaluate the proposed method through simulation-based experiments and comparative analysis against evolutionary baselines in order to assess convergence behavior, solution stability, and deployment efficiency.
\end{itemize}

The remainder of this paper is organized as follows. Section~\ref{sec:related} reviews the most relevant literature on UAV-assisted emergency communications and optimization-based deployment strategies. Section \ref{sec:problem} provides the system model and formulates our research problem. Section~\ref{sec:method} details the proposed HKQEA methodology. Section~\ref{sec:results} presents the experimental setup and the evaluation results. Finally, Section~\ref{sec:conclusion} concludes the paper and outlines directions for future research.

\section{Related Work} \label{sec:related}

The use of unmanned aerial vehicles (UAVs) as aerial base stations has received growing attention as a practical means of restoring wireless connectivity in disaster-affected areas \cite{b47}. From an optimization standpoint, post-disaster UAV deployment is challenging because the system must provide broad and reliable coverage under limited resources while respecting operational constraints such as separation, interference control, and deployment feasibility. Existing studies may be broadly grouped into three categories: exact optimization methods, heuristic and metaheuristic approaches, and quantum-inspired evolutionary methods.

Early studies have often relied on exact optimization formulations to model UAV deployment rigorously. In particular, constrained coverage-oriented formulations have been addressed through set-covering or related combinatorial models \cite{b48}. For example, Park \emph{et al.} \cite{b2} employed a Branch-and-Price (B\&P) framework to solve an emergency communication deployment problem with exact optimization guarantees. Similarly, mixed-integer linear programming (MILP) formulations have been used to optimize UAV positioning and maximize the number of served ground users, sometimes in conjunction with clustering-based decomposition strategies to improve tractability in larger instances \cite{b50}. Although such methods provide strong modeling fidelity and, in some cases, provable optimality, their computational cost grows rapidly with problem size, which limits their practical applicability in large-scale and time-sensitive disaster scenarios.

To improve scalability, a substantial body of work has turned to heuristic, metaheuristic, and artificial-intelligence-driven optimization approaches \cite{b51}. Survey studies such as \cite{b34} have highlighted the increasing role of these methods in UAV planning and control problems. Representative examples include Particle Swarm Optimization for deployment and monitoring tasks in disaster settings \cite{b27}, energy-aware bandit-based trajectory optimization \cite{b26}, and hybrid Ant Colony Optimization--Genetic Algorithm strategies for smart-city emergency response scenarios \cite{b28}. In the context of UAV-assisted communications, NSGA-II-based formulations have also been used to balance competing objectives such as fleet size and quality-of-service satisfaction \cite{b52}. These approaches generally offer improved search flexibility and better scalability than exact solvers; however, they often depend strongly on parameter settings, may converge prematurely in constrained search spaces, and frequently assume that the number of UAVs is fixed in advance.

Quantum-inspired evolutionary algorithms have been introduced to further strengthen global search capability and reduce stagnation in local optima. Foundational work by Han and Kim \cite{b35,b20} established the quantum genetic algorithm paradigm based on qubit-inspired probabilistic representations, which was later extended to multi-objective settings by Kim \emph{et al.} \cite{b36}. More recent studies have explored quantum-inspired methods for UAV-related optimization. Ballinas \emph{et al.} \cite{b29}, for example, proposed a hybrid quantum genetic algorithm with fuzzy adaptive rotation for three-dimensional UAV placement, while Panjavarnam \emph{et al.} \cite{b49} introduced a quantum-inspired bi-level neuro-swarm framework for UAV-based disaster management. These studies demonstrate the potential of quantum-inspired search mechanisms to improve convergence behavior and exploration quality. Nevertheless, prior work in this direction has typically focused on coverage maximization or positioning with a predetermined fleet size, rather than jointly addressing fleet-size minimization and deployment feasibility under multiple operational constraints.

Table~\ref{compo} summarizes representative studies most closely related to the present work.

\begin{table*}[htbp]
\centering
\caption{Representative optimization approaches for UAV deployment in disaster communication scenarios}
\label{compo}
\small
\begin{tabular}{p{1.8cm} p{3.5cm} p{5.2cm} p{5.2cm}}
\hline
\textbf{Reference} & \textbf{Technique} & \textbf{Strength} & \textbf{Main limitation} \\
\hline
\cite{b2} & B\&P & Exact optimization framework for constrained deployment with optimality guarantees & High computational burden, limiting scalability in large instances \\

\cite{b50} & MILP + K-means & Rigorous formulation with good solution quality; clustering improves tractability & Primarily maximizes served users and typically assumes a fixed UAV fleet size \\

\cite{b52} & NSGA-II & Multi-objective search balancing deployment-related objectives and service performance & Performance can depend strongly on discretization and may suffer from premature convergence \\

\cite{b29} & HQGA + fuzzy rotation & Introduces quantum-inspired operators to improve search behavior in UAV placement & Focuses mainly on coverage-oriented placement and assumes a fixed number of UAVs \\

\cite{b49} & Quantum-inspired bi-level neuro-swarm optimization & Demonstrates the promise of quantum-inspired search in UAV disaster-management settings & Does not directly target joint fleet minimization with explicit coverage and separation constraints \\
\hline
\end{tabular}
\end{table*}

Despite this progress, several methodological gaps remain. First, many existing studies treat the fleet size as a fixed input rather than as an optimization variable, even though resource minimization is central in emergency-response settings. Second, prior formulations often address coverage quality and deployment feasibility separately, rather than jointly enforcing coverage, overlap control, and minimum inter-UAV separation within a unified framework. Third, while quantum-inspired methods have shown promise, their use in coverage-driven UAV deployment remains relatively limited, particularly when structured population initialization and explicitly calibrated constraint handling are required.

Motivated by these limitations, the present work develops a hybrid K-means Quantum-Inspired Evolutionary Algorithm (HKQEA) for post-disaster UAV deployment. The proposed framework combines structured initialization, calibrated penalty-based feasibility management, and a non-elitist quantum-inspired search strategy to jointly address deployment efficiency and operational feasibility. In this way, the work aims to advance beyond fixed-fleet placement schemes toward a more resource-aware and constraint-consistent optimization framework for disaster communication restoration.

\section{System Model and Problem Formulation} \label{sec:problem}

We consider the static deployment stage of a post-disaster wireless restoration scenario in which unmanned aerial vehicles (UAVs) operate as aerial base stations to provide provisional connectivity to affected users. In large disaster environments, a single UAV is generally insufficient because of limited communication range, finite endurance, and the spatial dispersion of users. Coordinated multi-UAV deployment is therefore required to extend coverage, improve service continuity, and maintain operational robustness. In this work, we focus on determining an initial UAV deployment configuration that is both resource-efficient and operationally feasible for emergency communication support \cite{b57,b58}.

\begin{figure}[htbp]
    \centering
    \includegraphics[width=0.48\textwidth]{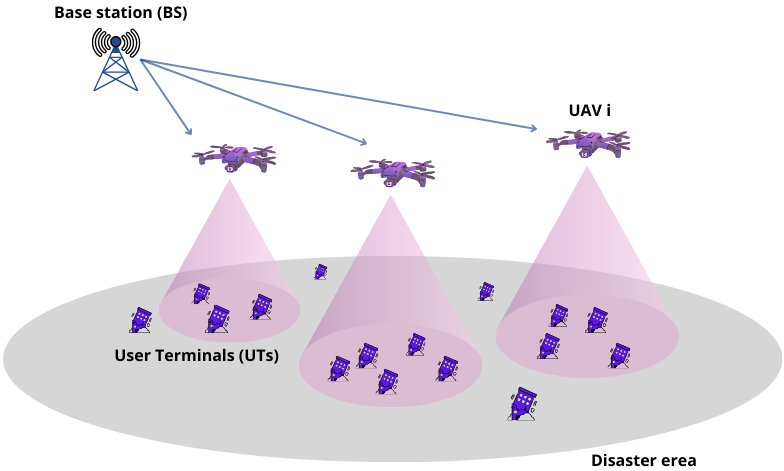}
    \caption{Illustration of the post-disaster UAV-assisted communication model.}
    \label{model}
\end{figure}

To model this scenario, we consider a two-dimensional area where the user terminals (UTs) are located. Each UAV is connected to the base station and covers the maximum number of UTs so they can send data to the base station. All UAVs have identical circular coverage areas with the same radius R. Figure~\ref{model} provides a conceptual illustration of this considered system model. Let the disaster region be represented by a two-dimensional area
\[
Z = [0,L_x]\times[0,L_y] \subset \mathbb{R}^2,
\]
where a set of user terminals (UTs)
\[
D=\{1,2,\dots,|D|\}
\]
must be served. Each UT $i\in D$ is located at known coordinates $(a_i,b_i)$, obtained for example from emergency reports, shelters, or responder databases. These UTs represent communication endpoints such as survivors' devices, first-responder radios, medical stations, or temporary relief points.

We assume that at most $N_{\max}$ UAVs are available for deployment. To model this, we define a set of potential UAV indices
\[
U=\{1,2,\dots,N_{\max}\},
\]
where each index $j\in U$ corresponds to a possible UAV slot. If UAV $j$ is activated, it is placed at continuous coordinates $(X_j,Y_j)\in Z$ and provides circular coverage of radius $R$. A binary variable $x_j$ indicates whether UAV $j$ is deployed, while a binary assignment variable $y_{ij}$ indicates whether UT $i$ is assigned to UAV $j$.

Accordingly, the decision variables are defined as
\begin{equation}
x_j=
\begin{cases}
1, & \text{if UAV } j \text{ is deployed},\\
0, & \text{otherwise},
\end{cases}
\qquad j\in U,
\end{equation}
and
\begin{equation}
y_{ij}=
\begin{cases}
1, & \text{if UT } i \text{ is assigned to UAV } j,\\
0, & \text{otherwise},
\end{cases}
\qquad i\in D,\; j\in U.
\end{equation}

The objective is to minimize the number of deployed UAVs while ensuring that every UT is served, that assignments are coverage-feasible, and that active UAVs satisfy a minimum separation requirement. In the present study, this separation threshold is denoted by $d_{\min}$ and is set to $2R$, consistent with the operational spacing assumption adopted in our experiments.

\subsection{Optimization Objective}

The fleet-size minimization objective is expressed as
\begin{equation}
\min \sum_{j\in U} x_j.
\end{equation}

This objective reflects the need to restore connectivity using the smallest feasible number of UAVs, thereby reducing deployment cost, logistical burden, and resource consumption.

\subsection{Constraints}

\subsubsection{Unique service assignment}
Each UT must be assigned to exactly one UAV:
\begin{equation}
\sum_{j\in U} y_{ij}=1,
\qquad \forall i\in D.
\end{equation}

This constraint ensures complete service to all UTs while avoiding redundant assignment in the optimization model.

\subsubsection{Deployment-assignment consistency}
A UT can be assigned only to an active UAV:
\begin{equation}
y_{ij}\leq x_j,
\qquad \forall i\in D,\; \forall j\in U.
\end{equation}

\subsubsection{Coverage feasibility}
If UT $i$ is assigned to UAV $j$, then UAV $j$ must lie within communication range $R$ of that UT. This conditional coverage requirement is enforced as
\begin{equation}
(a_i-X_j)^2+(b_i-Y_j)^2 \leq R^2 + M_c(1-y_{ij}),
\qquad \forall i\in D,\; \forall j\in U,
\end{equation}
where $M_c$ is a sufficiently large constant. When $y_{ij}=1$, the inequality reduces to the required coverage condition; when $y_{ij}=0$, the constraint is relaxed.

\subsubsection{Minimum inter-UAV separation}
To avoid unsafe proximity between simultaneously deployed UAVs, any two active UAVs must satisfy
\begin{equation}
(X_j-X_k)^2+(Y_j-Y_k)^2 \geq d_{\min}^2 - M_d(2-x_j-x_k),
\qquad \forall j,k\in U,\; j<k,
\end{equation}
where $M_d$ is a sufficiently large constant. Thus, the separation requirement is enforced only when both UAVs are active.

\subsubsection{Deployment-area bounds}

All UAV coordinates must remain within the disaster region:
\begin{equation}
0 \leq X_j \leq L_x,\qquad 0 \leq Y_j \leq L_y,
\qquad \forall j\in U.
\end{equation}

\subsubsection{Resource limit}
The number of active UAVs cannot exceed the available fleet:
\begin{equation}
\sum_{j\in U} x_j \leq N_{\max}.
\end{equation}

\subsection{Compact Formulation}

The complete optimization problem can therefore be written as

\begin{equation}
\begin{aligned}
\min \quad & \sum_{j\in U} x_j \\[2pt]
\text{s.t.}\quad
& \sum_{j\in U} y_{ij}=1,
&& \forall i\in D, \\
& y_{ij}\le x_j,
&& \forall i\in D,\ \forall j\in U, \\
& (a_i-X_j)^2+(b_i-Y_j)^2 \\
& \qquad \le R^2 + M_c(1-y_{ij}),
&& \forall i\in D,\ \forall j\in U, \\
& (X_j-X_k)^2+(Y_j-Y_k)^2 \\
& \qquad \ge d_{\min}^2 - M_d(2-x_j-x_k),
&& \forall j,k\in U,\ j<k, \\
& 0 \le X_j \le L_x,\quad 0 \le Y_j \le L_y,
&& \forall j\in U, \\
& \sum_{j\in U} x_j \le N_{\max}, \\
& x_j\in\{0,1\},
&& \forall j\in U, \\
& y_{ij}\in\{0,1\},
&& \forall i\in D,\ \forall j\in U.
\end{aligned}
\label{eq:placement_problem}
\end{equation}

The resulting problem is a constrained mixed-integer nonlinear optimization problem. It exhibits set-covering characteristics because the model seeks the smallest set of active UAVs whose coverage regions collectively serve all UTs, but differs from a classical discrete set covering problem in that UAV positions are optimized directly in continuous space. This joint activation-and-placement structure makes the problem computationally challenging, particularly as the number of UTs and the deployment area increase. Consequently, exact solution methods become difficult to scale, which motivates the use of advanced metaheuristic search strategies.

\section{Proposed HKQEA Methodology} \label{sec:method}

This section presents the proposed Hybrid K-means Quantum-Inspired Evolutionary Algorithm (HKQEA) for solving the joint UAV activation-and-placement problem defined in the previous section. The method is designed for the static deployment stage of post-disaster communication restoration, where the objective is to determine both \emph{how many} UAVs should be activated and \emph{where} they should be positioned in order to satisfy coverage and separation requirements with minimum resource usage.

Unlike the earlier discrete candidate-site interpretation, the proposed algorithm operates directly in continuous space and is therefore aligned with the formulation introduced in Section~\ref{sec:problem}. Each candidate solution simultaneously encodes UAV activation decisions and UAV coordinates. Constraint satisfaction is handled through a feasibility-aware bi-objective evaluation, while the search itself combines structured K-means-based initialization, non-elitist NSGA-II parent selection, and a quantum-inspired learning update implemented entirely on classical hardware.

\subsection{Design Rationale}

The optimization problem considered in this work is computationally challenging because it combines binary deployment decisions with continuous spatial variables under multiple operational constraints. As the number of user terminals (UTs) and the deployment budget increase, the search space grows rapidly, making exact solution methods difficult to scale for time-sensitive disaster-response settings. This motivates the use of population-based search methods.

Classical evolutionary and swarm-based optimizers have shown strong performance in deployment and coverage problems; however, their behavior can degrade in highly constrained settings where poor initialization, rapid loss of diversity, or overly aggressive elitism may lead to premature convergence. Quantum-inspired evolutionary algorithms reduce this risk by maintaining probabilistic states over multiple configurations simultaneously \cite{b20}; however, their standard formulations operate in the pre-measurement amplitude space and do not directly accommodate continuous spatial variables. Thus, the proposed HKQEA is designed to address these limitations through three complementary mechanisms:
\begin{itemize}
    \item \textbf{quantum-inspired probabilistic encoding}, to 
    represent binary deployment decisions as continuous activation 
    scores;
    \item \textbf{structured initialization}, to seed the search near spatially relevant user-density regions;
    \item \textbf{non-elitist generational replacement}, to preserve exploration pressure throughout the run;
    \item \textbf{quantum-inspired learning updates}, to guide candidate solutions toward promising regions without relying solely on stochastic variation.
\end{itemize}

The term \emph{quantum-inspired} is used here in the algorithmic sense: the method does not employ physical qubits or quantum hardware, but instead encodes binary deployment decisions as 
continuous activation scores in $[0,1]$, playing the role of Q-bit measurement probabilities~\cite{b20}, and adapts the directional bias of the quantum rotation gate to this post-collapse space by pulling each gene toward the local and global best solutions.
\subsection{Solution Representation and Decoding}

Let $N_{\max}$ denote the maximum number of UAVs that may be deployed. Each individual is encoded as a real-valued vector
\[
Q=
(s_1,\tilde{x}_1,\tilde{y}_1,\; s_2,\tilde{x}_2,\tilde{y}_2,\; \dots,\; s_{N_{\max}},\tilde{x}_{N_{\max}},\tilde{y}_{N_{\max}})
\in [0,1]^{3N_{\max}},
\]
where, for each UAV slot $j\in\{1,\dots,N_{\max}\}$:
\begin{itemize}
    \item $s_j\in[0,1]$ is an activation score,
    \item $\tilde{x}_j\in[0,1]$ and $\tilde{y}_j\in[0,1]$ are normalized spatial coordinates.
\end{itemize}
$s_j \in [0,1]$ is a probabilistic encoding inspired 
by the Q-bit measurement probability~\cite{b20}.

The chromosome is decoded as follows. The physical UAV coordinates are obtained by
\begin{equation}
X_j=L_x \tilde{x}_j,\qquad Y_j=L_y \tilde{y}_j,
\end{equation}
and the binary deployment decision is obtained from the activation score through a threshold rule
\begin{equation}
x_j=
\begin{cases}
1, & \text{if } s_j \ge \tau,\\
0, & \text{otherwise},
\end{cases}
\end{equation}
where $\tau\in(0,1)$ is a fixed activation threshold. This threshold acts as a measurement operation, collapsing 
the probabilistic activation score to a deterministic binary 
deployment decision $x_j \in \{0,1\}$.

Under this representation, the active UAV set associated with solution $Q$ is
\begin{equation}
A(Q)=\{j \in \{1,\dots,N_{\max}\}\;:\; x_j=1\}.
\end{equation}
This decoding step makes the evolutionary representation consistent with the optimization objective, since the number of active UAVs is computed directly as
\begin{equation}
f_1(Q)=\sum_{j=1}^{N_{\max}} x_j.
\end{equation}

\subsection{Assignment Induction During Fitness Evaluation}

The mathematical formulation uses binary assignment variables $y_{ij}$ to indicate service relationships between UTs and deployed UAVs. Rather than explicitly evolving these assignment variables, HKQEA derives them deterministically from the decoded deployment.

For a decoded solution $Q$, each UT $i$ is processed as follows:
\begin{enumerate}
    \item compute the set of active UAVs whose coverage radius contains the UT,
    \[
    \mathcal{A}_i(Q)=\left\{j\in A(Q): (a_i-X_j)^2+(b_i-Y_j)^2 \le R^2 \right\};
    \]
    \item if $\mathcal{A}_i(Q)\neq \emptyset$, assign UT $i$ to its nearest feasible active UAV,
    \[
    j^\star(i)=\arg\min_{j\in \mathcal{A}_i(Q)} d_{ij},
    \qquad 
    d_{ij}=\sqrt{(a_i-X_j)^2+(b_i-Y_j)^2};
    \]
    \item set
    \[
    y_{ij}=
    \begin{cases}
    1, & \text{if } j=j^\star(i),\\
    0, & \text{otherwise}.
    \end{cases}
    \]
\end{enumerate}

If $\mathcal{A}_i(Q)=\emptyset$, then UT $i$ is marked uncovered and incurs a feasibility penalty. This deterministic assignment rule enforces unique service assignment whenever coverage is available and makes the evaluation procedure directly consistent with the system model.

\subsection{K-means-Guided Sparse Initialization}

A high-quality initial population is important in constrained deployment problems. To generate spatially meaningful yet diverse initial individuals, HKQEA uses a K-means-guided sparse initialization procedure.

For each individual, an initial fleet size $k^{(n)}$ is first sampled from $\{1,2,\dots,N_{\max}\}$. K-means clustering is then applied to the UT coordinates using $k^{(n)}$ clusters:
\begin{equation}
\min_{C^{(n)}} \sum_{i=1}^{|D|} \min_{c \in C^{(n)}} \|(a_i,b_i)-c\|^2,
\end{equation}
where
\[
C^{(n)}=\{c^{(n)}_1,\dots,c^{(n)}_{k^{(n)}}\}
\]
denotes the centroid set for individual $n$.

The first $k^{(n)}$ UAV slots are initialized near these centroids after Gaussian perturbation:
\begin{equation}
\hat{c}^{(n)}_j=\operatorname{clip}\!\left(c^{(n)}_j+\epsilon_j,\;[0,L_x]\times[0,L_y]\right),
\qquad
\epsilon_j \sim \mathcal{N}(0,\sigma^2 I),
\end{equation}
and then normalized into $[0,1]^2$ to generate $\tilde{x}_j$ and $\tilde{y}_j$. Their corresponding activation scores are sampled above the threshold $\tau$ so that they begin as active UAVs. The remaining $N_{\max}-k^{(n)}$ slots are initialized with low activation scores and random coordinates in the search region.

This strategy serves two purposes. First, it places initially active UAVs near demand-concentrated regions, which improves early feasibility. Second, by varying $k^{(n)}$ across individuals and perturbing the centroids, it promotes diversity in both fleet size and spatial arrangement.

\subsection{Bi-objective Fitness Evaluation and Constraint Handling}

The original problem is a constrained single-objective optimization problem with the primary goal of minimizing the number of deployed UAVs. To enable feasibility-aware evolutionary search, HKQEA evaluates each solution through a bi-objective surrogate:
\begin{equation}
\min F(Q)=\big(f_1(Q),\,f_2(Q)\big),
\end{equation}
where $f_1(Q)$ is the fleet-size objective and $f_2(Q)$ quantifies infeasibility together with a secondary deployment-quality term.

\subsubsection{Primary objective}
The first objective counts the number of active UAVs:
\begin{equation}
f_1(Q)=\sum_{j=1}^{N_{\max}} x_j.
\end{equation}

\subsubsection{Feasibility and quality objective}
Let
\begin{equation}
N_{\mathrm{unc}}(Q)=\sum_{i\in D}\mathbf{1}\!\left[\mathcal{A}_i(Q)=\emptyset\right]
\end{equation}
denote the number of uncovered UTs, and let
\begin{equation}
V_{\mathrm{sep}}(Q)=\sum_{1\le j<k\le N_{\max}} x_j x_k \max\!\big(0,\, d_{\min}-d_{jk}\big)
\end{equation}
denote the cumulative minimum-separation violation, where
\begin{equation}
d_{jk}=\sqrt{(X_j-X_k)^2+(Y_j-Y_k)^2}.
\end{equation}

Among feasible or near-feasible configurations, we further use the aggregate service distance
\begin{equation}
T_{\mathrm{srv}}(Q)=\sum_{i\in D}\sum_{j=1}^{N_{\max}} y_{ij} d_{ij}
\end{equation}
as a secondary quality measure. The second objective is therefore defined as
\begin{equation}
f_2(Q)=
\lambda_1 N_{\mathrm{unc}}(Q)
+\lambda_2 V_{\mathrm{sep}}(Q)
+\lambda_3 T_{\mathrm{srv}}(Q),
\label{eq:f2_revised}
\end{equation}
where $\lambda_1,\lambda_2,\lambda_3>0$ are penalty coefficients.

The trade-off between ($f_1(Q)$) and ($f_2(Q)$) is handled through the bi-objective ranking procedure rather than through a single weighted aggregation of the two objectives. This is important because a direct scalarization could allow solutions with very small fleet sizes to dominate the search even when they leave many UTs uncovered or violate inter-UAV separation requirements. In the proposed formulation, ($f_1(Q)$) preserves the primary resource-efficiency objective, whereas ($f_2(Q)$) acts as a feasibility-aware deployment-quality objective. The largest penalty is assigned to uncovered UTs because coverage restoration is the primary operational requirement in the post-disaster scenario. Separation violations are penalized next to discourage unsafe or redundant UAV configurations, while the aggregate service distance is used only as a secondary discriminator among solutions with comparable feasibility. Therefore, the search first favors configurations that satisfy coverage and safety requirements, and then promotes smaller and more compact UAV deployments along the resulting Pareto front.

This formulation intentionally assigns the largest emphasis to uncovered UTs, followed by separation violations, while using service distance only as a secondary discriminator among solutions with comparable feasibility. In contrast to the earlier draft, no separate overlap-count penalty is introduced here: unique service assignment is enforced deterministically during evaluation, and geometric redundancy is already controlled by the minimum-separation constraint.

The penalty-based strategy is adopted because it is simple to integrate, computationally inexpensive, and well suited to the proposed non-elitist evolutionary framework. Its empirical calibration and comparison with alternative feasibility-handling mechanisms are reported in the experimental section.

\subsection{Selection, Variation, and Non-Elitist Replacement}

After all individuals are evaluated according to $\big(f_1(Q),f_2(Q)\big)$, the population is ranked using the standard NSGA-II non-dominated sorting procedure and crowding distance. In the proposed framework, however, this ranking is used \emph{only} for parent selection and not for elitist survival.

NSGA-II non-dominated sorting and crowding-distance ranking are adopted because the proposed formulation contains two competing objectives, fleet-size minimization and feasibility-aware deployment quality, for which Pareto-based ranking can preserve the trade-off without requiring scalarization weights or additional decomposition/indicator parameters.

Binary tournament selection is applied using dominance rank and crowding distance, thereby favoring individuals that achieve stronger trade-offs between fleet-size reduction and feasibility preservation. Selected parents then generate offspring through variation operators applied gene-wise to the encoded vector.

\subsubsection{Crossover}
Uniform crossover is used with probability $p_c$. For each gene, the offspring inherits the corresponding component from one of the two parents with equal probability. This allows both activation patterns and coordinate information to recombine efficiently.

\subsubsection{Mutation}
Gaussian mutation is applied independently to each gene with probability $p_m$. For activation-score genes and coordinate genes alike, the mutation rule is
\begin{equation}
q \leftarrow \operatorname{clip}(q+\delta,0,1),
\qquad
\delta\sim \mathcal{N}(0,\sigma_m^2).
\end{equation}
This introduces local perturbations while keeping all encoded values within the admissible interval.

\subsubsection{Non-elitist generational replacement}
Unlike standard NSGA-II, which selects survivors from the combined parent-offspring population, HKQEA uses a non-elitist generational policy in which the entire parent population is replaced by the offspring population after learning updates are applied. This choice is motivated by the need to preserve exploratory pressure in a highly constrained search space where overly strong elitism may lock the population into structurally similar deployment patterns too early.

\subsection{Quantum-Inspired Learning Update}

The quantum-inspired component of HKQEA is implemented as a directed learning operator acting on the encoded vector after crossover and mutation. Let $B^{(g)}$ denote the best individual in the current generation according to dominance rank, crowding distance, and then $f_2$ as a tie-breaker, and let $G^{(g)}$ denote the best solution found globally up to generation $g$.

For each offspring individual $Q$ and for each gene $q_\ell$ in its representation, the update is defined as
\begin{equation}
q_\ell \leftarrow \operatorname{clip}\!\left(
q_\ell + \eta\big(B^{(g)}_\ell-q_\ell\big)
      + \eta\big(G^{(g)}_\ell-q_\ell\big),
0,1\right),
\label{eq:q_update}
\end{equation}
where $\eta\in(0,1)$ is the learning rate.

This update is a continuous analogue of the quantum rotation gate~\cite{b20}: rather than rotating amplitude pairs $[\alpha_i, \beta_i]^{\top}$ in the pre-measurement Hilbert space, it replicates the directional bias of the rotation gate directly in the post-collapse probability space $[0,1]$, pulling each gene 
toward the activation patterns of the local best $B^{(g)}$ and global best $G^{(g)}$. The first term promotes local adaptation to the current search landscape, while the second introduces long-term memory and stabilizes global progress.
Importantly, the same update rule is applied consistently to both activation-score genes and coordinate genes. As a result, the learning mechanism simultaneously refines \emph{which} UAV slots should remain active and \emph{where} the active UAVs should move in the deployment region.

\subsection{Algorithm Summary}

\begin{figure}[H]
    \centering
    \includegraphics[scale=0.30]{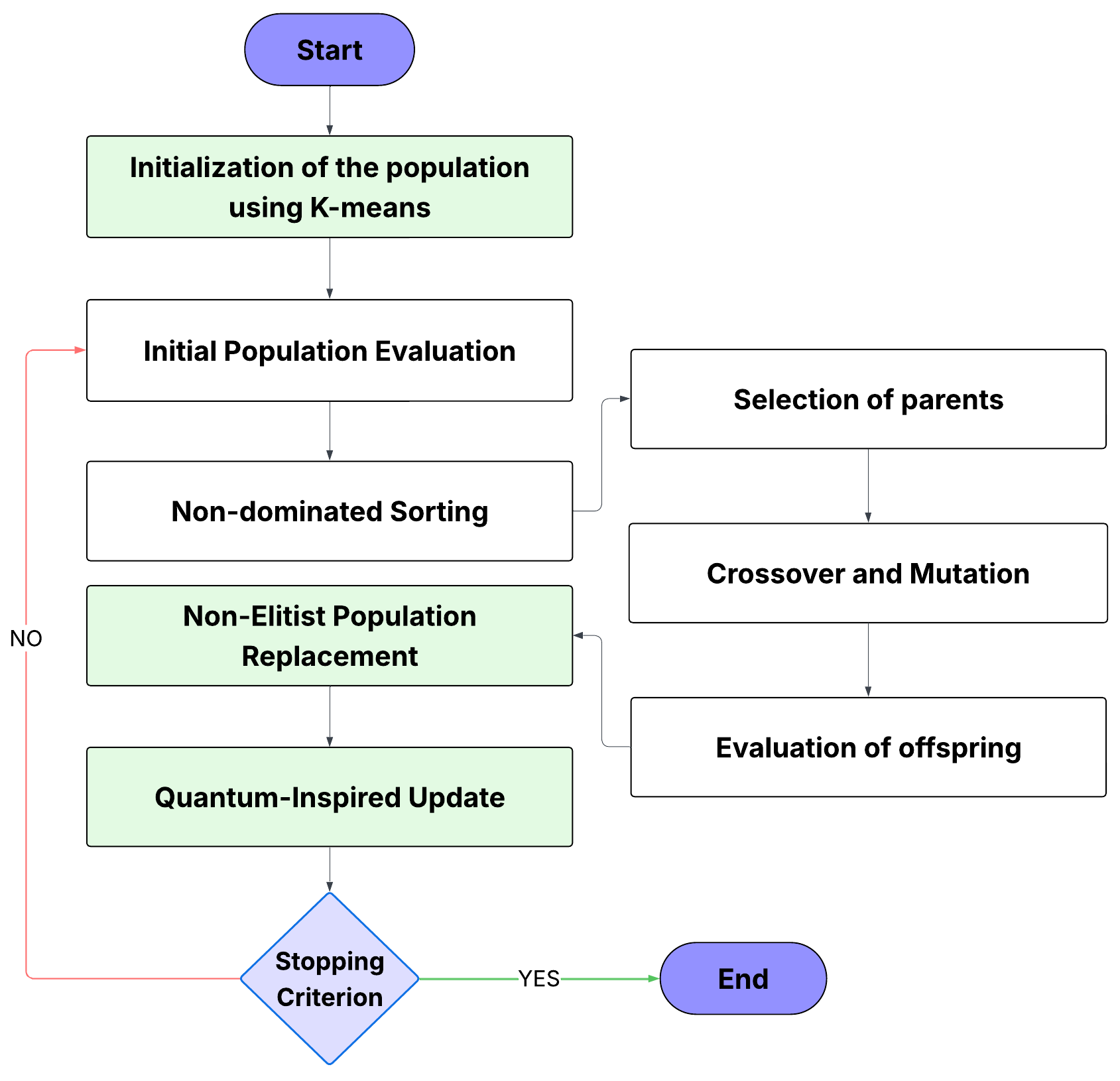}
    \caption{Algorithmic flow of the proposed HKQEA}
    \label{flow}
\end{figure}

The overall workflow of HKQEA is summarized in Algorithm~\ref{alg:hkqea_revised}. The procedure begins with K-means-guided sparse initialization, followed by repeated cycles of evaluation, non-dominated ranking, parent selection, crossover, mutation, quantum-inspired learning, and full generational replacement. Fig.~\ref{flow} shows the steps of the proposed algorithm. The K-means hybrid initialization, the non-elitist replacement strategy, and the quantum-inspired update are highlighted in green.

The final output is an approximation of the Pareto set induced by fleet-size minimization and feasibility-aware deployment quality.

\begin{algorithm}[htbp]
\footnotesize
\caption{Hybrid K-means Quantum-Inspired Evolutionary Algorithm (HKQEA)}
\label{alg:hkqea_revised}
\begin{algorithmic}[1]
\Statex \textbf{Input:} UT set $D$, deployment area $[0,L_x]\times[0,L_y]$, coverage radius $R$, maximum fleet size $N_{\max}$
\Statex \textbf{Parameters:} population size $N$, generations $G$, crossover probability $p_c$, mutation probability $p_m$, learning rate $\eta$, activation threshold $\tau$, penalty coefficients $(\lambda_1,\lambda_2,\lambda_3)$

\Statex \textbf{Initialization}
\For{$n=1$ to $N$}
    \State Sample initial fleet size $k^{(n)} \in \{1,\dots,N_{\max}\}$
    \State Run K-means on UT coordinates with $k^{(n)}$ clusters
    \State Initialize $k^{(n)}$ UAV slots near the centroids with activation scores above $\tau$
    \State Initialize remaining slots with low activation scores and random coordinates
    \State Construct encoded individual $Q^{(n)} \in [0,1]^{3N_{\max}}$
\EndFor
\State $P^{(0)} \gets \{Q^{(1)},\dots,Q^{(N)}\}$
\State Initialize global best solution $G^{(0)}$

\Statex \textbf{Main evolutionary loop}
\For{$g=0$ to $G-1$}
    \ForAll{$Q \in P^{(g)}$}
        \State Decode activation scores and coordinates to obtain $\{x_j,X_j,Y_j\}_{j=1}^{N_{\max}}$
        \State Induce UT assignments $\{y_{ij}\}$ by nearest feasible active-UAV assignment
        \State Evaluate $f_1(Q)$ and $f_2(Q)$
    \EndFor
    \State Perform non-dominated sorting and crowding-distance computation on $P^{(g)}$
    \State Select mating pool by binary tournament selection
    \State Generate offspring population $\hat{P}^{(g)}$ using uniform crossover and Gaussian mutation
    \State Identify current-generation best individual $B^{(g)}$
    \ForAll{$Q \in \hat{P}^{(g)}$}
        \For{each gene $q_\ell$ in $Q$}
            \State Update $q_\ell$ according to \eqref{eq:q_update}
        \EndFor
    \EndFor
    \State Set $P^{(g+1)} \gets \hat{P}^{(g)}$ \Comment{non-elitist replacement}
    \State Update global best solution $G^{(g+1)}$
\EndFor

\Statex \textbf{Output:} Final non-dominated solution set and corresponding UAV deployment configurations
\end{algorithmic}
\end{algorithm}

\subsection{Computational Complexity and Scalability Analysis}
\label{subsec:complexity}

We analyze the computational complexity of HKQEA in terms of the number of user terminals $|\mathcal{D}|$, the maximum number of UAV slots $N_{\max}$, the population size $P$, the number of generations $G$, and the average number of K-means iterations $I_{\mathrm{km}}$. Each individual contains $3N_{\max}$ real-valued genes, representing one activation score and two normalized spatial coordinates for each UAV slot.

The K-means-guided initialization is performed once. For one individual, its worst-case cost is $O(I_{\mathrm{km}}|\mathcal{D}|N_{\max})$, and for the full population it becomes $T_{\mathrm{init}}=O(P I_{\mathrm{km}}|\mathcal{D}|N_{\max})$. During each generation, evaluating one individual requires decoding the UAV activation and coordinate variables, assigning UTs to feasible UAVs, computing coverage and service-distance terms, and checking pairwise UAV-separation violations. These steps require $O(N_{\max})$ for decoding, $O(|\mathcal{D}|N_{\max})$ for UT-to-UAV assignment and coverage evaluation, and $O(N_{\max}^{2})$ for pairwise separation checking. Therefore, the fitness-evaluation cost for one individual is $T_{\mathrm{eval}}^{(1)}=O(|\mathcal{D}|N_{\max}+N_{\max}^{2})$, and for the full population it is $T_{\mathrm{eval}}^{(P)}=O(P(|\mathcal{D}|N_{\max}+N_{\max}^{2}))$.

The NSGA-II ranking step contributes $O(P^{2})$ due to non-dominated sorting, while crowding-distance computation adds $O(P\log P)$ and is dominated by the sorting term. The crossover, mutation, and quantum-inspired learning update operate gene-wise on vectors of length $3N_{\max}$, giving $O(PN_{\max})$ per generation. Thus, the quantum-inspired update remains linear in the encoded solution size and does not introduce exponential or quantum-hardware overhead.

Combining these terms, the per-generation complexity is $T_{\mathrm{gen}}=O(P(|\mathcal{D}|N_{\max}+N_{\max}^{2})+P^{2}+PN_{\max})$. Since $PN_{\max}$ is typically dominated by the evaluation cost, the overall per-run complexity can be written as $T_{\mathrm{HKQEA}}=O(P I_{\mathrm{km}}|\mathcal{D}|N_{\max}+G[P(|\mathcal{D}|N_{\max}+N_{\max}^{2})+P^{2}])$.

This analysis shows that, for a fixed UAV budget, the main evaluation cost scales approximately linearly with the number of UTs. However, increasing $N_{\max}$ introduces a quadratic term because all UAV pairs must be checked for minimum-separation feasibility, and the $P^{2}$ NSGA-II sorting term may become significant for very large populations. Therefore, HKQEA is realistic for moderate-to-large disaster scenarios when the UAV budget is bounded, but larger deployments would benefit from parallel fitness evaluation, spatial indexing or grid-based neighbor search, adaptive population sizing, and regional decomposition of very large affected areas.

\section{Experimental Evaluation and Results} \label{sec:results}

This section presents the experimental setup, evaluation criteria, calibration of the constraint-handling mechanism, and the comparative performance of the proposed HKQEA against relevant evolutionary baselines. All reported results are obtained under the continuous UAV activation-and-placement formulation introduced earlier and are therefore consistent with the representation and decoding scheme adopted in the proposed methodology.

\subsection{Experimental Setup}


All experiments were conducted over a square deployment region $Z=[0,100]\times[0,100]$, where the coordinates are expressed in normalized distance units. Specifically, one coordinate unit corresponds to $U=100~\mathrm{m}$. Thus, the simulated deployment area corresponds to a physical region of $(100U)\times(100U)=10~\mathrm{km}\times10~\mathrm{km}$, which represents a post-disaster urban district with disrupted ground communication infrastructure~\cite{Gao2023rapid}. The deployment region contains (|D|=32) fixed user terminals (UTs). The UAVs are modeled as low-altitude platforms (LAPs) acting as aerial base stations. Based on the adopted air-to-ground propagation model, UAVs operating at altitudes between ($500~\mathrm{m}$) and ($1000~\mathrm{m}$) can provide an effective coverage radius of approximately ($2~\mathrm{km}$) in suburban environments~\cite{AlHourani2014optimal}. 

Accordingly, the coverage radius used in the simulation is $R=20U=2~\mathrm{km}$. The minimum inter-UAV separation is set to $d_{\min}=2R=4~\mathrm{km}$, which serves as a conservative operational spacing constraint to reduce unsafe proximity, excessive geometric redundancy, and potential inter-UAV interference. Therefore, the reported value (R=20) should be interpreted as (20) normalized distance units, not as ($20~\mathrm{m}$) or ($20~\mathrm{km}$). The maximum available fleet size is set to ($N_{\max}=10$). Under the representation introduced in Section~\ref{sec:method}, each candidate solution consists of ($3N_{\max}=30$) encoded decision variables: one activation score and two normalized spatial coordinates for each UAV slot.


The proposed HKQEA was implemented in Python using the DEAP framework. Unless otherwise stated, the population size was set to $N=100$, the maximum number of generations was set to $G=1000$, the crossover and mutation probabilities were fixed at $p_c=0.7$ and $p_m=0.2$, respectively, and the learning-rate parameter of the quantum-inspired update was set to $\eta=0.2$. A fixed activation threshold $\tau$ was used to decode activation scores into binary deployment decisions. Each experiment was repeated over $50$ independent runs to reduce sensitivity to stochastic effects and to enable descriptive statistical comparison across methods.

The K-means-guided sparse initialization strategy described in Section~\ref{sec:method} was used to generate the initial population. All simulations were executed on a system equipped with an Intel\textsuperscript{\textregistered} Core\texttrademark{} i7-7600U CPU @ 2.80~GHz and 16~GB of RAM.

For methodological fairness, all compared evolutionary methods used the same deployment encoding, decoding, objective evaluation, and stopping criterion. The differences between them are therefore restricted to the search dynamics: HKQEA employs both the quantum-inspired learning update and non-elitist generational replacement, the elitist HKQEA retains the same learning mechanism but uses standard elitist survival, and the NSGA-II baseline uses the same encoding and evaluation pipeline without the quantum-inspired learning step.

The present evaluation is designed as a controlled proof-of-concept study rather than an exhaustive scalability benchmark. For each candidate solution, decoding the activation and coordinate variables requires ($O(N_{\max}$)), assigning UTs to feasible active UAVs and computing coverage-related quantities requires ($O(|D|N_{\max})$), and evaluating pairwise minimum-separation violations requires ($O(N_{\max}^{2})$). Thus, for a population of size $(P)$ evolved over $(G)$ generations, the dominant per-run computational complexity is
\begin{equation}
\mathcal{O}!\left(
P I_{\mathrm{km}} |D| N_{\max}
+
G\left[
P\left(|D|N_{\max}+N_{\max}^{2}\right)
+
P^{2}
\right]
\right),
\label{eq:hkqea-complexity}
\end{equation}

where $(P)$ denotes the population size, $(G)$ is the number of generations, $(I_{\mathrm{km}})$ is the number of K-means iterations, $(|D|)$ is the number of UTs, and $(N_{\max})$ is the maximum UAV fleet size. This indicates that, for a fixed UAV budget, the main deployment-evaluation cost scales approximately linearly with the number of UTs, whereas increasing ($N_{\max}$) introduces an additional quadratic term due to pairwise UAV-separation checks. A broader empirical scalability study involving larger UT populations, heterogeneous spatial distributions, and varying UAV budgets is left for future work.

\subsection{Evaluation Metrics}

Performance is assessed using metrics that reflect deployment efficiency, coverage quality, feasibility, and computational cost.

\subsubsection{Fleet size}
The primary optimization outcome is the number of active UAVs in the decoded final solution:
\begin{equation}
S^\ast=\sum_{j=1}^{N_{\max}} x_j.   
\end{equation}

This metric directly reflects the objective of minimizing the deployed fleet size.

\subsubsection{Coverage percentage}
Coverage performance is measured as the proportion of UTs that lie within the coverage region of at least one active UAV:
\begin{equation}
 \mathrm{Co}(\%)=
\left(
\frac{N_{\mathrm{covered}}}{N_{\mathrm{total}}}
\right)\times 100.   
\end{equation}

\subsubsection{Non-overlap percentage}
Because service assignment is enforced uniquely during decoding, overlap is evaluated here as a \emph{geometric redundancy} metric rather than an assignment ambiguity metric. Let $N_{\mathrm{single}}$ denote the number of covered UTs that fall inside the coverage region of exactly one active UAV. The non-overlap percentage is then defined as
\begin{equation}
\mathrm{Over}(\%)=
\left(
\frac{N_{\mathrm{single}}}{N_{\mathrm{covered}}}
\right)\times 100.
\end{equation}
This measure quantifies the extent to which the deployment avoids redundant geometric coverage.

\subsubsection{Minimum-distance satisfaction}
Let $N_{\mathrm{pairs}}=\binom{S^\ast}{2}$ denote the number of active UAV pairs, and let $N_{\mathrm{viol}}$ denote the number of those pairs that violate the required minimum separation distance $d_{\min}=2R$. The distance-satisfaction metric is defined as
\begin{equation}
\mathrm{Dis}(\%)=
\left(
1-\frac{N_{\mathrm{viol}}}{N_{\mathrm{pairs}}}
\right)\times 100,
\end{equation}
with $\mathrm{Dis}(\%)=100$ when fewer than two UAVs are active.

\subsubsection{Execution time}
Computational efficiency is measured by the total runtime, in seconds, required for one full optimization run.

Together, these metrics provide a balanced view of how effectively a method minimizes fleet size while preserving full or near-full service coverage, limiting redundant overlap, and satisfying operational safety constraints.

\subsection{Calibration of the Constraint-Handling Strategy}

Before evaluating the overall optimization performance of HKQEA, we first examined the effect of the penalty coefficients used in the second objective and compared the selected penalty model against several alternative feasibility-handling mechanisms. The purpose of this study was not to claim universal superiority of one constraint-handling strategy, but rather to identify a practically effective configuration for the deployment problem considered here.

Table~\ref{tab:penalty_evaluation} reports the outcomes obtained for several candidate penalty settings. The results indicate that excessively small penalties lead to poor coverage preservation, whereas overly aggressive penalty weights do not further improve feasibility and may reduce search flexibility. Among the tested configurations, $(\lambda_1,\lambda_2,\lambda_3)=(1750,150,50)$ provided the best overall compromise, achieving full coverage, full non-overlap, and full separation satisfaction while maintaining a competitive fleet size.

\begin{table}[htbp]
\footnotesize
\centering
\caption{Calibration of penalty coefficients for the feasibility objective.}
\label{tab:penalty_evaluation}
\begin{tabular}{c c c c c}
\hline
\textbf{$(\lambda_1,\lambda_2,\lambda_3)$} & \textbf{$S^\ast$} & \textbf{Co (\%)} & \textbf{Over (\%)} & \textbf{Dis (\%)} \\
\hline
(1,1,1)             & 6 & 31.25 & 90.00 & 99.60 \\
(100,10,5)          & 7 & 90.62 & 100.00 & 99.00 \\
(250,20,10)         & 8 & 96.87 & 90.40 & 99.00 \\
(500,50,20)         & 6 & 84.37 & 100.00 & 100.00 \\
(750,70,30)         & 8 & 93.75 & 100.00 & 98.80 \\
(1000,100,30)       & 8 & 96.87 & 100.00 & 100.00 \\
(1250,130,30)       & 8 & 96.87 & 100.00 & 99.00 \\
(1500,130,50)       & 7 & 96.87 & 100.00 & 99.80 \\
(1750,150,50)       & 8 & 100.00 & 100.00 & 100.00 \\
(2000,150,50)       & 8 & 96.87 & 100.00 & 99.40 \\
\hline
\end{tabular}
\end{table}

\begin{figure}[htbp]
    \centering
    \includegraphics[scale=0.4]{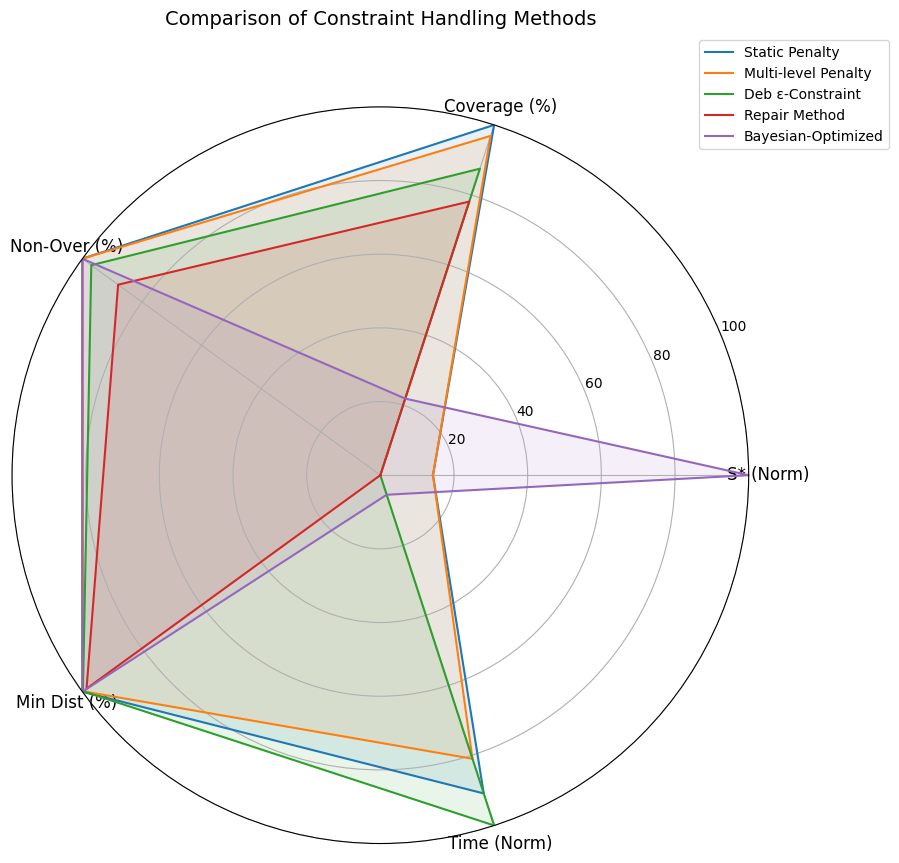}
    \caption{Normalized comparison of the evaluated constraint-handling methods.}
    \label{radar}
\end{figure}

Using this calibrated setting, we compared the adopted static-penalty formulation with four alternative feasibility-handling approaches: a multi-level penalty scheme, Deb's $\varepsilon$-constraint method, a repair-based strategy, and a Bayesian-optimized penalty (BOP) model. As shown in Table~\ref{tab:methods}, the calibrated static-penalty formulation provides the most balanced result across coverage preservation, overlap control, separation satisfaction, and runtime. Figure~\ref{radar} offers a normalized visual summary of these trade-offs.

\begin{table}[htbp]
\scriptsize
\centering
\caption{Comparison of constraint-handling methods under the same evolutionary search framework.}
\label{tab:methods}
\begin{tabular}{l c c c c c}
\toprule
\textbf{Method} & \textbf{$S^\ast$} & \textbf{Co (\%)} & \textbf{Over (\%)} & \textbf{Dis (\%)} & \textbf{Time (s)} \\
\midrule
Static penalty             & 8 & 100.00 & 100.00 & 100.00 & 161 \\
Multi-level penalty        & 7 & 90.62  & 96.55  & 99.39  & 293 \\
Deb $\epsilon$-constraint  & 9 & 87.50  & 97.00  & 99.80  & 202 \\
Repair method              & 9 & 78.12  & 88.00  & 98.60  & 1404 \\
BOP & 2 & 21.87  & 100.00 & 100.00 & 1337 \\
\bottomrule
\end{tabular}
\end{table}

\subsection{Performance of the Proposed HKQEA}

Table~\ref{tab:metrics_summary} summarizes the descriptive statistics of HKQEA over 50 independent runs. The proposed method achieved a best fully feasible deployment of $8$ UAVs with $100\%$ coverage, $100\%$ non-overlap, and $100\%$ minimum-distance satisfaction. Across all runs, HKQEA maintained a mean fleet size of $8.20$ UAVs, an average coverage rate of $98.94\%$, and near-perfect non-overlap and separation satisfaction, while exhibiting relatively low variability in all feasibility-related metrics.

\begin{table}[htbp]
\scriptsize
\centering
\caption{Descriptive statistics of HKQEA over 50 independent runs.}
\label{tab:metrics_summary}
\begin{tabular}{c c c c c c c}
\hline
\textbf{Method} & \textbf{Statistic} & \textbf{$S^\ast$} & \textbf{Co (\%)} & \textbf{Over (\%)} & \textbf{Dis (\%)} & \textbf{Time (s)} \\
\hline
\multirow{4}{*}{\textbf{HKQEA}}
& Best  & 8.00 & 100.00 & 100.00 & 100.00 & 81.93 \\
& Avg   & 8.20 & 98.94  & 99.94  & 99.68  & 148.61 \\
& Worst & 9.00 & 87.50  & 96.77  & 98.99  & 298.93 \\
& Std   & 0.40 & 2.22   & 0.45   & 0.32   & 55.20 \\
\hline
\end{tabular}
\end{table}

\begin{figure}[htbp]
\centering
\begin{minipage}{.48\textwidth}
  \centering
  \includegraphics[width=\textwidth]{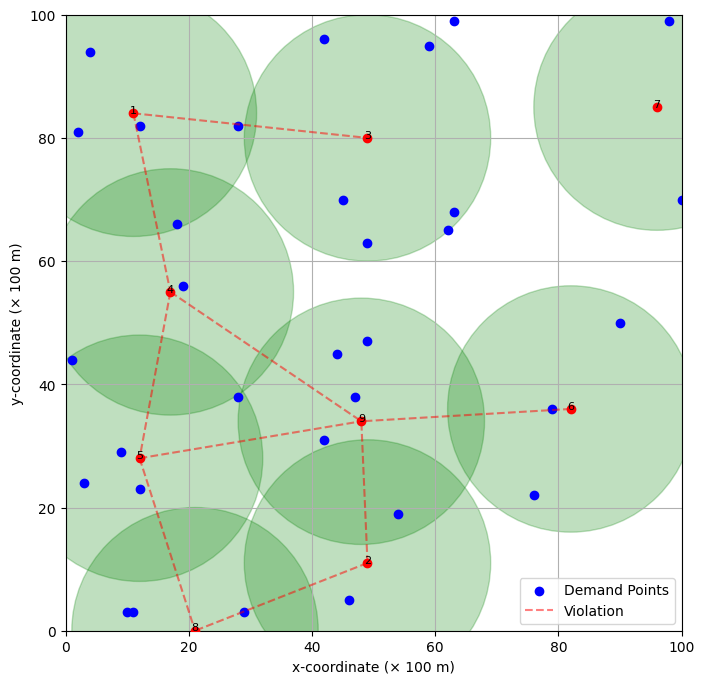}
  \caption{HKQEA - Run 21, initial generation.}
  \label{fig:test1}
\end{minipage}\hfill
\begin{minipage}{.48\textwidth}
  \centering
  \includegraphics[width=\textwidth]{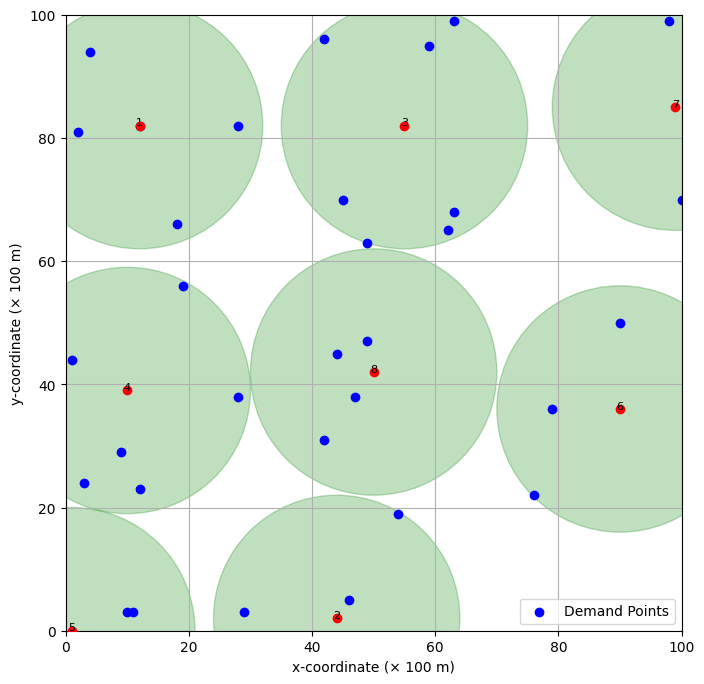}
  \caption{HKQEA - Run 21, final deployment.}
  \label{fig:test2}
\end{minipage}
\end{figure}

The best fully feasible run was obtained in Run~21. Figures~\ref{fig:test1} and \ref{fig:test2} illustrate the corresponding initial and final spatial configurations. The initial population, produced by the K-means-guided initialization process, starts from demand-aware yet diverse candidate deployments, whereas the final population reflects the progressive refinement of active UAV locations toward a compact and feasible solution.

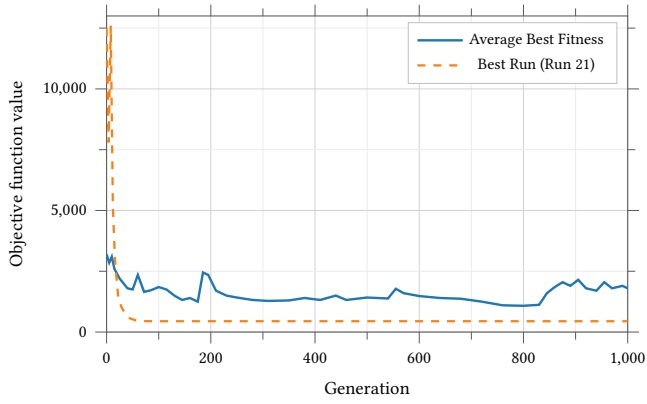
\begin{figure}[h]
\centering
\begin{tikzpicture}
\begin{axis}[
    width=\columnwidth,
    height=0.68\columnwidth,
    xmin=0, xmax=1000,
    ymin=0, ymax=13000,
    xlabel={Generation},
    ylabel={Objective function value},
    xlabel style={font=\footnotesize},
    ylabel style={font=\footnotesize},
    tick label style={font=\scriptsize},
    legend style={
        font=\scriptsize,
        at={(0.98,0.98)},
        anchor=north east,
        draw=black!25,
        fill=white,
        line width=0.3pt
    },
    grid=both,
    major grid style={black!18, line width=0.3pt},
    minor grid style={black!8, line width=0.2pt},
    minor tick num=1,
    tick align=outside,
    axis line style={black!65, line width=0.4pt},
    tick style={black!65, line width=0.35pt},
    scaled ticks=false,
    every axis plot/.append style={line width=0.9pt},
]

\addplot[
    color=ieeeblue,
    solid,
    mark=none
] coordinates {
    (0,3200)   (5,2850)   (10,3100)  (15,2600)  (25,2200)
    (40,1800)  (50,1750)  (60,2350)  (72,1650)  (85,1720)
    (100,1850) (115,1750) (130,1500) (145,1320) (160,1400)
    (175,1250) (185,2450) (195,2350) (210,1700) (230,1500)
    (250,1420) (280,1320) (310,1280) (350,1300) (380,1400)
    (410,1320) (440,1500) (460,1320) (500,1420) (540,1380)
    (555,1780) (570,1600) (600,1480) (640,1400) (680,1370)
    (720,1250) (760,1100) (800,1080) (830,1120) (845,1600)
    (860,1850) (875,2050) (890,1900) (905,2150) (920,1800)
    (940,1700) (955,2050) (970,1800) (990,1900) (1000,1800)
};

\addplot[
    color=ieeeorange,
    dashed,
    mark=none,
    line width=0.85pt
] coordinates {
    (0,12500) (4,7800)  (8,12600) (12,5200) (16,3000)
    (22,1500) (28,1050) (35,780)  (45,560)  (60,450)
    (100,445) (200,445) (300,445) (400,445) (500,445)
    (600,445) (700,445) (800,445) (900,445) (1000,445)
};

\legend{Average Best Fitness, Best Run (Run 21)}

\end{axis}
\end{tikzpicture}
\caption{HKQEA-Run 21: Convergence behavior of the proposed optimization procedure over 1000 generations.}
\label{convergence}
\end{figure}

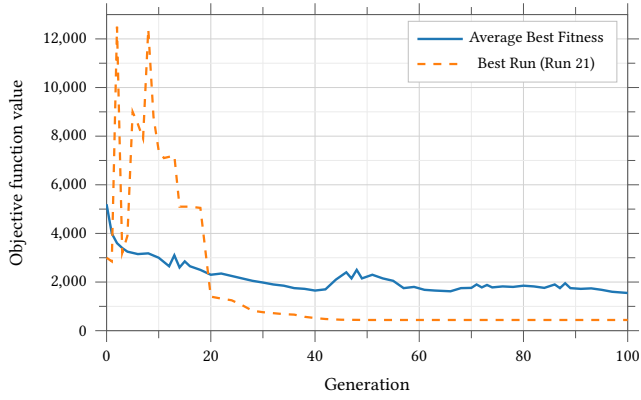
\begin{figure}[h]
\centering
\begin{tikzpicture}
\begin{axis}[
    width=\columnwidth,
    height=0.68\columnwidth,
    xmin=0, xmax=100,
    ymin=0, ymax=13000,
    xlabel={Generation},
    ylabel={Objective function value},
    xlabel style={font=\footnotesize},
    ylabel style={font=\footnotesize},
    tick label style={font=\scriptsize},
    xtick={0,20,40,60,80,100},
    ytick={0,2000,4000,6000,8000,10000,12000},
    legend style={
        font=\scriptsize,
        at={(0.98,0.98)},
        anchor=north east,
        draw=black!25,
        fill=white,
        line width=0.3pt
    },
    grid=both,
    major grid style={black!18, line width=0.3pt},
    minor grid style={black!8, line width=0.2pt},
    minor tick num=1,
    tick align=outside,
    axis line style={black!65, line width=0.4pt},
    tick style={black!65, line width=0.35pt},
    scaled ticks=false,
    every axis plot/.append style={line width=0.9pt},
]

\addplot[
    color=ieeeblue,
    solid,
    mark=none
] coordinates {
    (0,5200)  (1,4000)  (2,3600)  (3,3400)  (4,3250)
    (6,3150)  (8,3180)  (10,3000) (12,2650) (13,3100)
    (14,2600) (15,2850) (16,2650) (18,2500) (20,2300)
    (22,2350) (24,2250) (26,2150) (28,2050) (30,1980)
    (32,1900) (34,1850) (36,1750) (38,1720) (40,1650)
    (42,1700) (44,2100) (46,2400) (47,2150) (48,2500)
    (49,2150) (51,2300) (53,2150) (55,2050) (57,1750)
    (59,1800) (61,1680) (63,1650) (66,1620) (68,1750)
    (70,1760) (71,1900) (72,1780) (73,1880) (74,1780)
    (76,1820) (78,1800) (80,1850) (82,1820) (84,1760)
    (86,1900) (87,1750) (88,1950) (89,1750) (91,1720)
    (93,1740) (95,1680) (97,1600) (100,1550)
};

\addplot[
    color=ieeeorange,
    dashed,
    mark=none,
    line width=0.85pt
] coordinates {
    (0,3000)  (1,2850)  (2,12500) (3,3200)  (4,3900)
    (5,9000)  (6,8500)  (7,7900)  (8,12400) (9,8800)
    (10,7400) (11,7100) (13,7200) (14,5100) (16,5100)
    (18,5050) (20,1400) (22,1320) (24,1250) (26,1050)
    (28,820)  (30,760)  (32,720)  (34,680)  (36,660)
    (38,570)  (40,520)  (42,480)  (45,450)  (50,440)
    (55,440)  (60,440)  (65,440)  (70,440)  (75,440)
    (80,440)  (85,440)  (90,440)  (95,440)  (100,440)
};

\legend{Average Best Fitness, Best Run (Run 21)}

\end{axis}
\end{tikzpicture}
\caption{HKQEA-Run 21: early-stage convergence over the first 100 generations.}
\label{fig:placeholder}
\end{figure}

The convergence behavior of this run is shown in Figures~\ref{convergence} and \ref{fig:placeholder}. The solution quality improves rapidly during the early generations, with near-optimal performance emerging after approximately 40 generations. Thereafter, the search enters a stable regime in which no further material improvements are observed.


\subsection{Comparative Analysis Against Baseline Methods}

To assess the contribution of the proposed design choices, HKQEA was compared against three baselines: a standard NSGA-II implementation, an elitist variant of HKQEA, and Particle Swarm Optimization (PSO). The purpose of this comparison is threefold. First, NSGA-II serves as a widely used multi-objective evolutionary benchmark. Second, the elitist HKQEA isolates the effect of replacing the proposed non-elitist survival policy with conventional elitist replacement while keeping the remaining components of the method unchanged. Third, PSO serves as a classical swarm-based metaheuristic benchmark, allowing the quantum-inspired learning update to be evaluated against a structurally related but conceptually distinct directional update mechanism.

Table~\ref{tab:comparison} reports descriptive statistics over 50 runs for the three methods. Importantly, the entries in the \emph{Best} row are \emph{marginal} best values for each metric across all runs and therefore do not necessarily correspond to a single jointly feasible solution. For this reason, Table~\ref{tab:best_run} separately reports the best fully feasible run identified for each method.

\begin{table}[htbp]
\scriptsize
\centering
\caption{Descriptive statistics over 50 runs for the compared methods. The \emph{Best}, \emph{Avg}, and \emph{Worst} entries are marginal statistics per metric and do not necessarily correspond to the same run.}
\label{tab:comparison}
\begin{tabular}{c c c c c c c}
\hline
\textbf{Method} & \textbf{Statistic} & \textbf{$S^\ast$} & \textbf{Co (\%)} & \textbf{Over (\%)} & \textbf{Dis (\%)} & \textbf{Time (s)} \\
\hline
\multirow{4}{*}{\textbf{HKQEA}}
& Best  & 8.00 & 100.00 & 100.00 & 100.00 & 81.93 \\
& Avg   & 8.20 & 98.94  & 99.94  & 99.68  & 148.61 \\
& Worst & 9.00 & 87.50  & 96.77  & 98.99  & 298.93 \\
& Std   & 0.40 & 2.22   & 0.45   & 0.32   & 55.20 \\
\hline

\multirow{4}{*}{\shortstack{\textbf{HKQEA}\\\textbf{(Elitist)}}}
& Best  & 9.00 & 100.00 & 100.00 & 99.60 & 124.69 \\
& Avg   & 9.00 & 100.00 & 100.00 & 99.20 & 156.68 \\
& Worst & 9.00 & 100.00 & 100.00 & 98.79 & 434.26 \\
& Std   & 0.40 & 0.00   & 0.00   & 0.18  & 45.13 \\
\hline

\multirow{4}{*}{\textbf{NSGA-II}}
& Best  & 6.00 & 100.00 & 100.00 & 100.00 & 85.74 \\
& Avg   & 7.34 & 89.38  & 98.87  & 99.58  & 100.07 \\
& Worst & 9.00 & 71.88  & 79.17  & 98.59  & 108.62 \\
& Std   & 0.68 & 6.64   & 3.24   & 0.26   & 5.36 \\
\hline
\multirow{4}{*}{\textbf{PSO}}
& Best  & 5.00 & 96.88 & 100.00 & 100.00 & 43.45 \\
& Avg   & 6.66 & 88.69  & 99.79  & 99.74  & 58.15 \\
& Worst & 8.00 & 68.75  & 96.43  & 98.59  & 69.58 \\
& Std   & 0.84 & 5.79   & 0.83   & 0.32   & 6.68 \\
\hline
\end{tabular}
\end{table}

The best fully feasible runs of the four methods are summarized in Table~\ref{tab:best_run}. HKQEA achieved a fully feasible deployment with 8 UAVs in Run~21 and converged within approximately 30--40 generations (Conv.Gen). PSO reached its best feasible solution with 8 UAVs within a comparable 30--40 generation range. The elitist HKQEA required 9 UAVs in its best feasible run and converged more slowly, while NSGA-II also reached a fully feasible 8-UAV solution but only after substantially more generations.

\begin{table}[htbp]
\scriptsize
\centering
\caption{Best fully feasible run obtained by each algorithm.}
\label{tab:best_run}
\begin{tabular}{c c c c c c c}
\hline
\textbf{Method} & \textbf{Run} & \textbf{$S^\ast$} & \textbf{Co (\%)} & \textbf{Over (\%)} & \textbf{Dis (\%)} & \textbf{Conv. Gen} \\
\hline
HKQEA            & 21 & 8 & 100.00 & 100.00 & 100.00 & 30--40 \\
Elitist HKQEA    & 42 & 9 & 100.00 & 100.00 & 98.99  & 80--100 \\
NSGA-II          & 40 & 8 & 100.00 & 100.00 & 100.00 & 100--150 \\
PSO          & 47 & 8 & 96.88 & 100.00 & 99.80 & 30--40 \\
\hline
\end{tabular}
\end{table}

\begin{figure}[htbp]
\centering
\begin{minipage}{.45\textwidth}
  \centering
  \includegraphics[width=\textwidth]{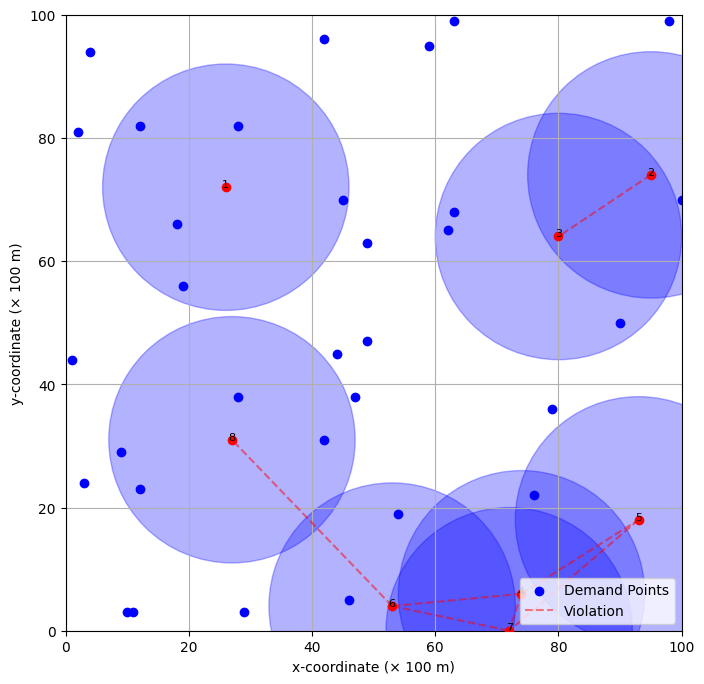}
  \caption{NSGA-II - Run 40, initial generation.}
  \label{0BLUE}
\end{minipage}\hfill
\begin{minipage}{.45\textwidth}
  \centering
  \includegraphics[width=.98\textwidth]{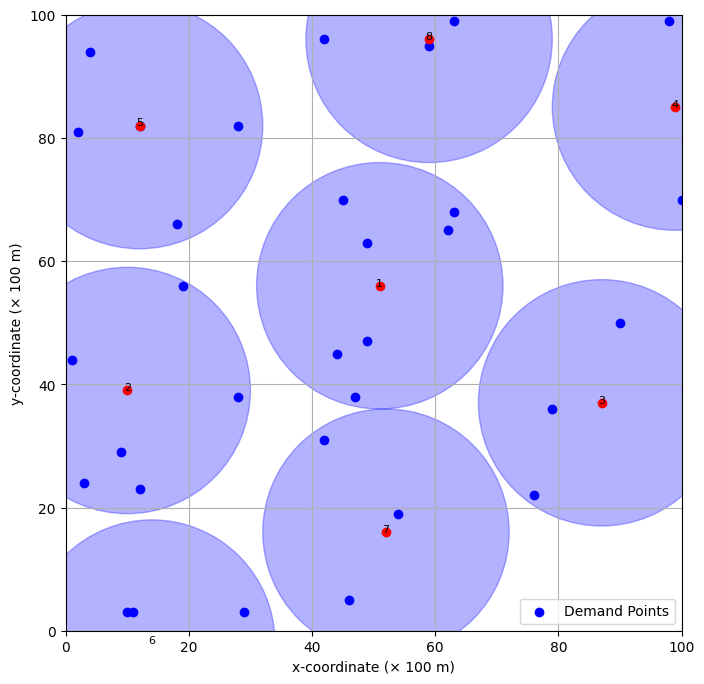}
  \caption{NSGA-II - Run 40, final deployment.}
  \label{1000BLUE}
\end{minipage}
\end{figure}

\begin{figure}[htbp]
\centering
\begin{minipage}{.45\textwidth}
  \centering
  \includegraphics[width=\textwidth]{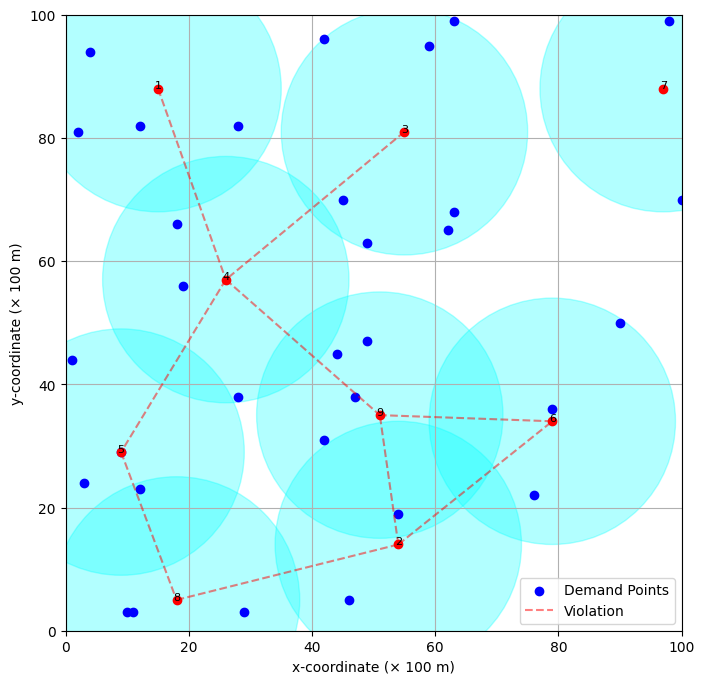}
  \caption{Elitist HKQEA - Run 42, initial generation.}
  \label{0CIAN}
\end{minipage}\hfill
\begin{minipage}{.45\textwidth}
  \centering
  \includegraphics[width=\textwidth]{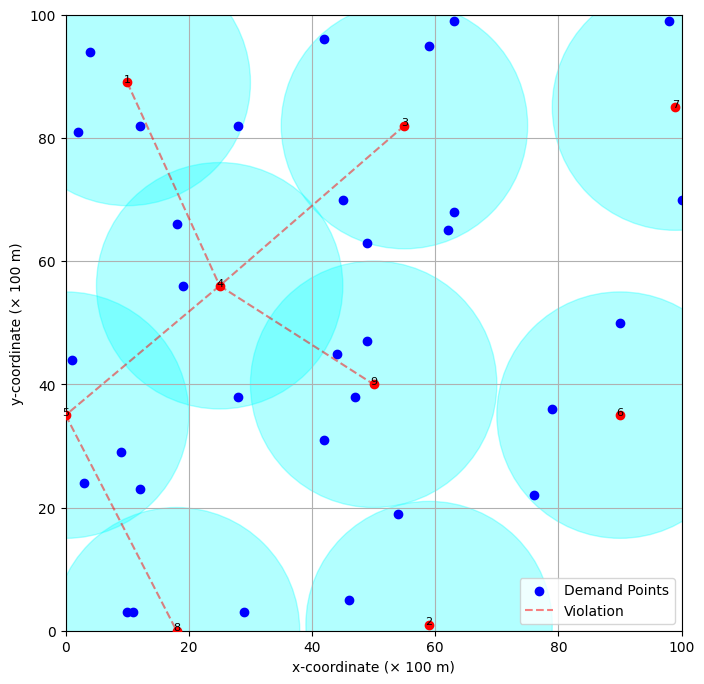}
  \caption{Elitist HKQEA - Run 42, final deployment.}
  \label{1000CIAN}
\end{minipage}
\end{figure}

\begin{figure}[htbp]
\centering
\begin{minipage}{.45\textwidth}
  \centering
  \includegraphics[width=\textwidth]{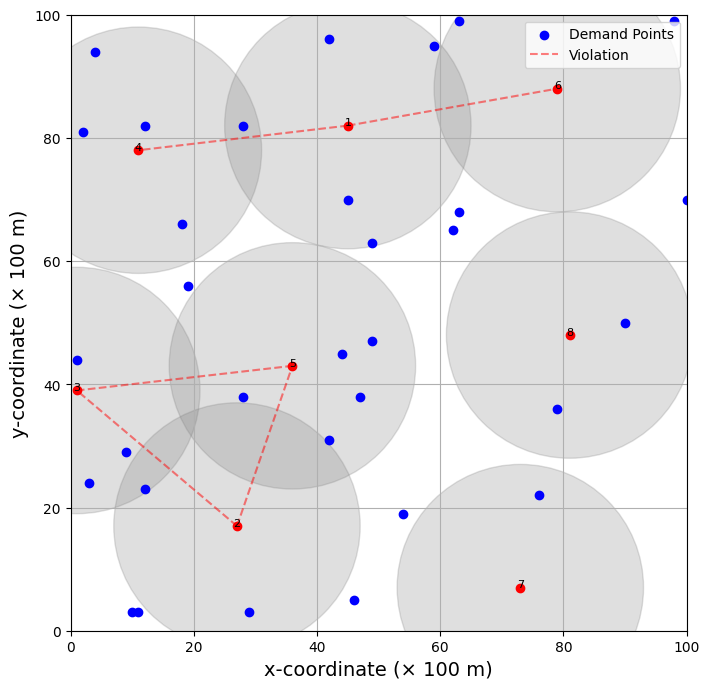}
  \caption{PSO - Run 47, initial generation.}
  \label{0GRAY}
\end{minipage}\hfill
\begin{minipage}{.45\textwidth}
  \centering
  \includegraphics[width=\textwidth]{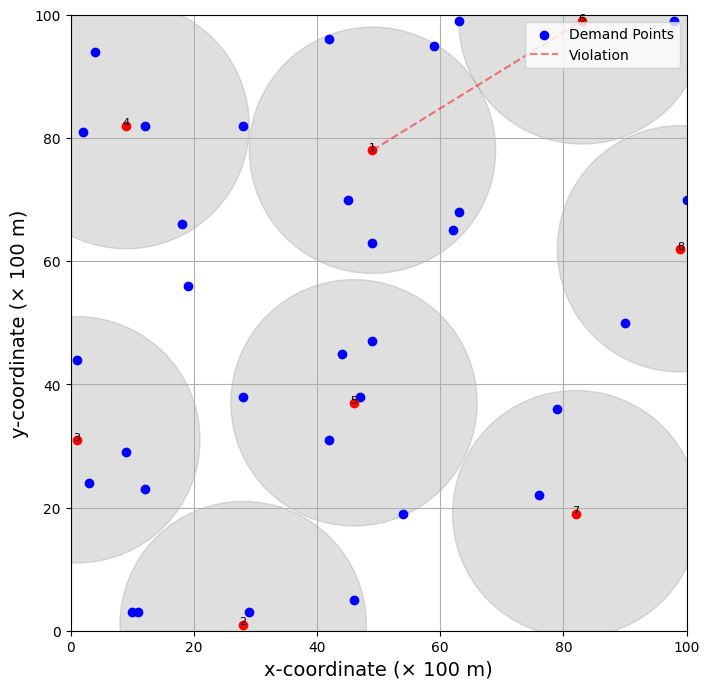}
  \caption{PSO - Run 47, final deployment.}
  \label{1000GRAY}
\end{minipage}
\end{figure}

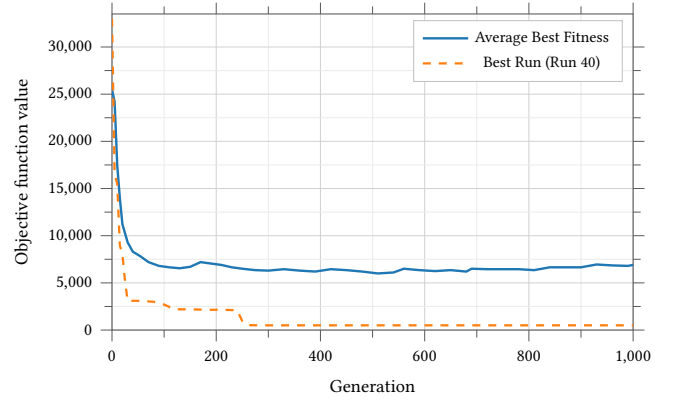
\begin{figure}[h]
\centering
\begin{tikzpicture}
\begin{axis}[
    width=\columnwidth,
    height=0.68\columnwidth,
    xmin=0, xmax=1000,
    ymin=0, ymax=33500,
    xlabel={Generation},
    ylabel={Objective function value},
    xlabel style={font=\footnotesize},
    ylabel style={font=\footnotesize},
    tick label style={font=\scriptsize},
    xtick={0,200,400,600,800,1000},
    ytick={0,5000,10000,15000,20000,25000,30000},
    legend style={
        font=\scriptsize,
        at={(0.98,0.98)},
        anchor=north east,
        draw=black!25,
        fill=white,
        line width=0.3pt
    },
    grid=both,
    major grid style={black!18, line width=0.3pt},
    minor grid style={black!8, line width=0.2pt},
    minor tick num=1,
    tick align=outside,
    axis line style={black!65, line width=0.4pt},
    tick style={black!65, line width=0.35pt},
    scaled ticks=false,
    every axis plot/.append style={line width=0.9pt},
]

\addplot[
    color=ieeeblue,
    solid,
    mark=none
] coordinates {
    (0,25500)  (5,24200)  (10,17500) (15,14000) (20,11200)
    (30,9300)  (40,8300)  (55,7800)  (70,7200)  (90,6800)
    (110,6650) (130,6550) (150,6700) (170,7200) (190,7050)
    (210,6900) (230,6650) (250,6500) (275,6350) (300,6300)
    (330,6450) (360,6300) (390,6200) (420,6450) (450,6350)
    (480,6200) (510,6000) (540,6100) (560,6500) (590,6350)
    (620,6250) (650,6350) (680,6200) (690,6500) (720,6450)
    (750,6450) (780,6450) (810,6350) (840,6650) (870,6650)
    (900,6650) (930,6950) (960,6850) (990,6800) (1000,6900)
};

\addplot[
    color=ieeeorange,
    dashed,
    mark=none,
    line width=0.85pt
] coordinates {
    (0,33000)  (5,16500)  (10,15500) (15,9000)  (20,8000)
    (25,5200)  (30,3100)  (50,3100)  (80,3000)  (100,2700)
    (120,2200) (150,2200) (180,2150) (210,2150) (240,2100)
    (250,950)  (260,520)  (300,500)  (350,500)  (400,500)
    (450,500)  (500,500)  (550,500)  (600,500)  (650,500)
    (700,500)  (750,500)  (800,500)  (850,500)  (900,500)
    (950,500)  (1000,500)
};

\legend{Average Best Fitness, Best Run (Run 40)}

\end{axis}
\end{tikzpicture}
\caption{NSGA-II - Run 40: convergence history.}
\label{NSGA}
\end{figure}

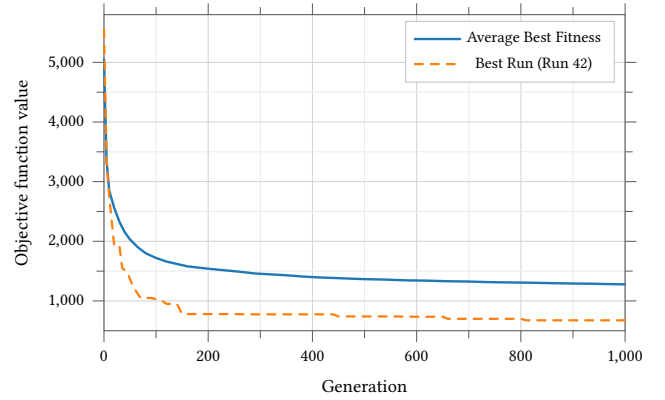
\begin{figure}[h]
\centering
\begin{tikzpicture}
\begin{axis}[
    width=\columnwidth,
    height=0.68\columnwidth,
    xmin=0, xmax=1000,
    ymin=500, ymax=5800,
    xlabel={Generation},
    ylabel={Objective function value},
    xlabel style={font=\footnotesize},
    ylabel style={font=\footnotesize},
    tick label style={font=\scriptsize},
    xtick={0,200,400,600,800,1000},
    ytick={1000,2000,3000,4000,5000},
    legend style={
        font=\scriptsize,
        at={(0.98,0.98)},
        anchor=north east,
        draw=black!25,
        fill=white,
        line width=0.3pt
    },
    grid=both,
    major grid style={black!18, line width=0.3pt},
    minor grid style={black!8, line width=0.2pt},
    minor tick num=1,
    tick align=outside,
    axis line style={black!65, line width=0.4pt},
    tick style={black!65, line width=0.35pt},
    scaled ticks=false,
    every axis plot/.append style={
        line width=0.9pt,
        line join=round,
        line cap=round
    },
]

\addplot[
    color=ieeeblue,
    solid,
    mark=none
] coordinates {
    (0,5050)   (5,3350)   (10,2850)  (20,2550)  (30,2320)
    (40,2150)  (50,2030)  (65,1900)  (80,1800)  (100,1720)
    (120,1660) (140,1620) (160,1580) (180,1560) (200,1540)
    (230,1515) (260,1490) (290,1460) (320,1445) (350,1430)
    (380,1410) (410,1395) (440,1385) (470,1375) (500,1365)
    (540,1358) (580,1345) (620,1340) (660,1330) (700,1325)
    (740,1315) (780,1310) (820,1305) (860,1298) (900,1292)
    (940,1288) (970,1282) (1000,1278)
};

\addplot[
    color=ieeeorange,
    dashed,
    mark=none,
    line width=0.85pt
] coordinates {
    (0,5550)   (5,3350)   (10,2750)  (20,1900)  (30,1900)
    (35,1550)  (45,1480)  (55,1250)  (70,1050)  (90,1050)
    (110,1010) (120,950)  (140,950)  (150,780)  (170,780)
    (200,780)  (240,780)  (280,775)  (320,775)  (360,775)
    (400,775)  (440,775)  (450,740)  (500,740)  (550,740)
    (600,735)  (650,735)  (660,700)  (700,700)  (750,700)
    (800,700)  (810,675)  (850,675)  (900,675)  (950,675)
    (1000,675)
};

\legend{Average Best Fitness, Best Run (Run 42)}

\end{axis}
\end{tikzpicture}
\caption{Elitist HKQEA - Run 42: convergence history.}
\label{elitist}
\end{figure}

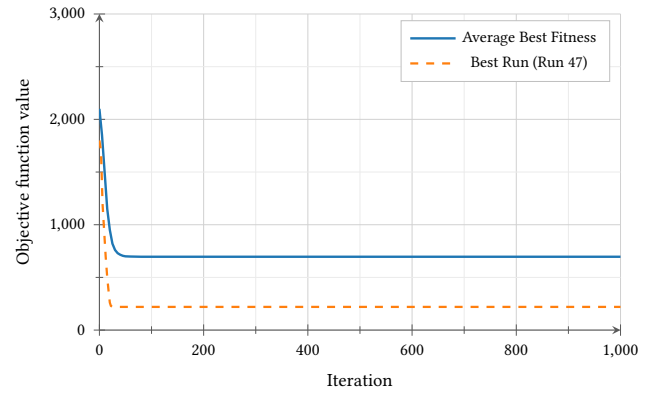
\begin{figure}[h]
\centering
\begin{tikzpicture}
\begin{axis}[
    width=\columnwidth,
    height=0.68\columnwidth,
    xmin=0, xmax=1000,
    ymin=0, ymax=3000,
    xlabel={Iteration},
    ylabel={Objective function value},
    xlabel style={font=\footnotesize},
    ylabel style={font=\footnotesize},
    tick label style={font=\scriptsize},
    xtick={0,200,400,600,800,1000},
    ytick={0,1000,2000,3000},
    legend style={
        font=\scriptsize,
        at={(0.98,0.98)},
        anchor=north east,
        draw=black!25,
        fill=white,
        line width=0.3pt
    },
    grid=both,
    major grid style={black!18, line width=0.3pt},
    minor grid style={black!8, line width=0.2pt},
    minor tick num=1,
    tick align=outside,
    axis line style={black!65, line width=0.4pt},
    tick style={black!65, line width=0.35pt},
    scaled ticks=false,
    every axis plot/.append style={line width=0.9pt},
    enlarge x limits=false,
    enlarge y limits=false,
    axis lines=left,
]
\addplot[
    color=ieeeblue,
    solid,
    mark=none
] coordinates {
    (0,2100)   (5,1850)   (10,1500)  (15,1150)  (20,950)
    (25,820)   (30,760)   (35,730)   (40,715)   (45,705)
    (50,700)   (60,698)   (70,697)   (80,696)   (90,696)
    (100,696)  (150,696)  (200,696)  (250,696)  (300,696)
    (350,696)  (400,696)  (450,696)  (500,696)  (550,696)
    (600,696)  (650,696)  (700,696)  (750,696)  (800,696)
    (850,696)  (900,696)  (950,696)  (1000,696)
};
\addplot[
    color=ieeeorange,
    dashed,
    mark=none,
    line width=0.85pt
] coordinates {
    (0,1800)   (3,1700)   (6,1200)   (9,950)    (12,700)
    (15,500)   (18,350)   (20,260)   (22,230)   (25,220)
    (30,220)   (40,220)   (50,220)   (60,220)   (70,220)
    (80,220)   (100,220)  (150,220)  (200,220)  (300,220)
    (400,220)  (500,220)  (600,220)  (700,220)  (800,220)
    (900,220)  (1000,220)
};
\legend{Average Best Fitness, Best Run (Run 47)}
\end{axis}
\end{tikzpicture}
\caption{PSO convergence over 1000 iterations.}
\label{pso}
\end{figure}

Figures~\ref{0BLUE}--\ref{pso} further illustrate the qualitative deployment patterns and convergence behavior of the two baseline methods.

From Table~\ref{tab:comparison} and Table~\ref{tab:best_run}, several observations can be made. First, the proposed HKQEA offers the strongest balance between fleet-size reduction and reliable feasibility preservation. Its average performance is consistently high, and the variability of its coverage, overlap, and separation metrics remains low across runs. Second, the elitist HKQEA exhibits stronger run-to-run regularity, but this stability is achieved at the cost of structural flexibility: once the population converges to a 9-UAV configuration, elitist survival makes it difficult to discover leaner feasible deployments. Third, although NSGA-II attains strong marginal metric values in some runs, its average behavior is substantially less stable, especially in terms of coverage preservation.  Fourth, PSO converges quickly, but this reflects premature commitment to small fleet sizes rather than genuine feasibility, its average behavior is substantially less stable and it never attains full coverage despite near-perfect non-overlap and minimum-separation satisfaction.\par

A particularly important observation concerns the apparent 5-UAV \emph{Best} value reported for PSO and  6-UAV \emph{Best} value reported for NSGA-II in Table~\ref{tab:comparison}. This value is a marginal statistic and does not correspond to the same fully feasible deployment reflected in the perfect metric values reported elsewhere in that row. When the comparison is restricted to the best \emph{jointly feasible} run for each algorithm, as shown in Table~\ref{tab:best_run}, PSO and NSGA-II requires 8 UAVs, matching HKQEA in fleet size but converging more slowly and with higher variability across runs for NSGA-II while PSO converges prematurely and never attains full coverage.

Overall, the results indicate that the proposed non-elitist HKQEA provides the most favorable trade-off among exploration capability, convergence speed, and reliable feasibility preservation for the studied deployment instance.

\subsection{Practical Resource Implications}

Although the primary contribution of this work is algorithmic rather than economic, the optimized fleet size has a direct practical implication for emergency deployment planning. In the studied scenario, HKQEA identifies a fully feasible deployment using (8) UAVs instead of the maximum available fleet of (10), corresponding to a (20\%) reduction in deployed hardware while preserving full coverage and separation satisfaction. For a representative industrial UAV platform with a unit procurement cost of approximately (\$13{,}000)~\cite{b54}, this reduction would correspond to an illustrative hardware-procurement saving of about (\$26{,}000) for the studied deployment instance. This estimate is intentionally limited to direct hardware count and should not be interpreted as a full economic model. A comprehensive operational cost analysis would require additional modeling of energy consumption, battery lifecycle, maintenance, charging or replacement logistics, and multi-trip deployment constraints, which are left for future work.

\section{Conclusion} \label{sec:conclusion}

This paper presented a hybrid quantum-inspired evolutionary framework for the static deployment of UAV-based aerial base stations in post-disaster communication scenarios. The problem was formulated as a joint activation-and-placement optimization task in continuous space, with the objective of minimizing the number of deployed UAVs while satisfying coverage and minimum-separation constraints. To address this problem, we developed the proposed Hybrid K-means Quantum-Inspired Evolutionary Algorithm (HKQEA), which combines K-means-guided initialization, a calibrated penalty-based feasibility objective, non-elitist evolutionary search, and a quantum-inspired learning update. Experimental results on the studied deployment instance showed that HKQEA achieved a best fully feasible solution with 8 UAVs, while maintaining 100\% coverage, 100\% non-overlap, and 100\% minimum-distance satisfaction. Across 50 independent runs, the method also exhibited strong average performance, low variability, and rapid early convergence.

Comparative evaluation against standard NSGA-II , PSO and an elitist variant of HKQEA further showed that the proposed non-elitist design provides a more favorable balance between exploration, convergence speed, and reliable feasibility preservation in constrained deployment problems. While the present study is limited to a static scenario with simplified geometric assumptions, the results demonstrate the promise of the proposed framework for resource-efficient post-disaster communication restoration. Future work will extend the model to broader deployment scenarios, dynamic UAV mobility, energy-aware operation, and more realistic communication and environmental constraints.

\section*{Ethics:}
The authors confirm that there are no ethical concerns associated with this study.

\section*{Conflicts of Interest:}
The authors declare no conflict of interest.

\section*{Data availibility:}
The datasets used and/or analysed during the current study available from the corresponding author on reasonable request.

\section*{Funding statement:}
This research did not receive a specific grant from any funding agency in the public, commercial or non-profit sectors.





\bibliographystyle{ieeetr} 
\bibliography{references}



\end{document}